%% file: ms.tex
\documentclass[sigconf]{acmart}

\usepackage{subfiles}
\AtBeginDocument{%
  }

\setcopyright{acmcopyright}
\copyrightyear{2018}
\acmYear{2018}
\acmDOI{XXXXXXX.XXXXXXX}

\acmConference[Conference acronym 'XX]{Make sure to enter the correct
  conference title from your rights confirmation emai}{June 03--05,
  2018}{Woodstock, NY}
\acmPrice{15.00}
\acmISBN{978-1-4503-XXXX-X/18/06}

\copyrightyear{2024}
\acmYear{2024}
\setcopyright{rightsretained}
\acmConference[HRI '24]{Proceedings of the 2024 ACM/IEEE International Conference on Human-Robot Interaction}{March 11--14, 2024}{Boulder, CO, USA}
\acmBooktitle{Proceedings of the 2024 ACM/IEEE International Conference on Human-Robot Interaction (HRI '24), March 11--14, 2024, Boulder, CO, USA}
\acmDOI{10.1145/3610977.3634999}
\acmISBN{979-8-4007-0322-5/24/03}

\usepackage[capitalise, nameinlink]{cleveref}

\usepackage{xspace}
\usepackage{fontawesome}
\newcommand{\fullname}[0]{Generative Expressive Motion\xspace}
\newcommand{\abv}[0]{GenEM\xspace}

\newcommand{\oracle}[0]{\textcolor[RGB]{125, 125, 125}{oracle animator}\xspace}
\newcommand{\genem}[0]{\textcolor[RGB]{255, 160, 139}{GenEM}\xspace}
\newcommand{\genemp}[0]{\textcolor[RGB]{255, 73, 34}{GenEM++}\xspace}

\begin{document}

\title{Generative Expressive Robot Behaviors \\ using Large Language Models}


\author{Karthik Mahadevan}
\affiliation{%
  \institution{Google Deepmind}
  \country{}
  }
\email{karthikm@dgp.toronto.edu}

\author{Jonathan Chien}
\affiliation{%
  \institution{Google Deepmind}
  \country{}
  }
\email{chienj@google.com}

\author{Noah Brown}
\affiliation{%
  \institution{Google Deepmind}
  \country{}
  }
\email{noahbrown@google.com}

\author{Zhuo Xu}
\affiliation{%
  \institution{Google Deepmind}
  \country{}
  }
\email{zhuoxu@google.com}

\author{Carolina Parada}
\affiliation{%
  \institution{Google Deepmind}
  \country{}
  }
\email{carolinap@google.com}

\author{Fei Xia}
\affiliation{%
  \institution{Google Deepmind}
  \country{}
  }
\email{xiafei@google.com}

\author{Andy Zeng}
\affiliation{%
  \institution{Google Deepmind}
  \country{}
  }
\email{andyzeng@google.com}

\author{Leila Takayama}
\affiliation{%
  \institution{Hoku Labs}
  \country{}
  }
\email{takayama@hokulabs.com}

\author{Dorsa Sadigh}
\affiliation{%
  \institution{Google Deepmind}
  \country{}
  }
\email{dorsas@google.com}
\renewcommand{\shortauthors}{Karthik Mahadevan et al.}

\input{Camera-Ready/Sections/Abstract}



\maketitle
\input{Camera-Ready/Sections/Introduction}
\input{Camera-Ready/Sections/Related_Work}
\input{Camera-Ready/Sections/Approach}
\input{Camera-Ready/Sections/User_Studies}
\input{Camera-Ready/Sections/Experiments}

\input{Camera-Ready/Sections/Discussion}
\input{Camera-Ready/Sections/Acknowledgements}
\bibliographystyle{ACM-Reference-Format}
\bibliography{references}

\end{document}

%% file: Camera-Ready/Sections/Abstract.tex
\begin{abstract}
    People employ expressive behaviors to effectively communicate and coordinate their actions with others, such as nodding to acknowledge a person glancing at them or saying \emph{``excuse me''} to pass people in a busy corridor. We would like robots to also demonstrate expressive behaviors in human-robot interaction. Prior work proposes rule-based methods that struggle to scale to new communication modalities or social situations, while data-driven methods require specialized datasets for each social situation the robot is used in. We propose to leverage the rich social context available from large language models (LLMs) and their ability to generate motion based on instructions or user preferences, to generate \emph{expressive robot motion} that is \textit{adaptable} and \textit{composable}, building upon each other. Our approach utilizes few-shot chain-of-thought prompting to translate human language instructions into parametrized control code using the robot's available and learned skills. Through user studies and simulation experiments, we demonstrate that our approach produces behaviors that users found to be competent and easy to understand. Supplementary material can be found at \url{https://generative-expressive-motion.github.io/}.

\end{abstract}

%% file: Camera-Ready/Sections/Introduction.tex
\begin{figure}
  \centering
  \includegraphics[width=1\columnwidth]{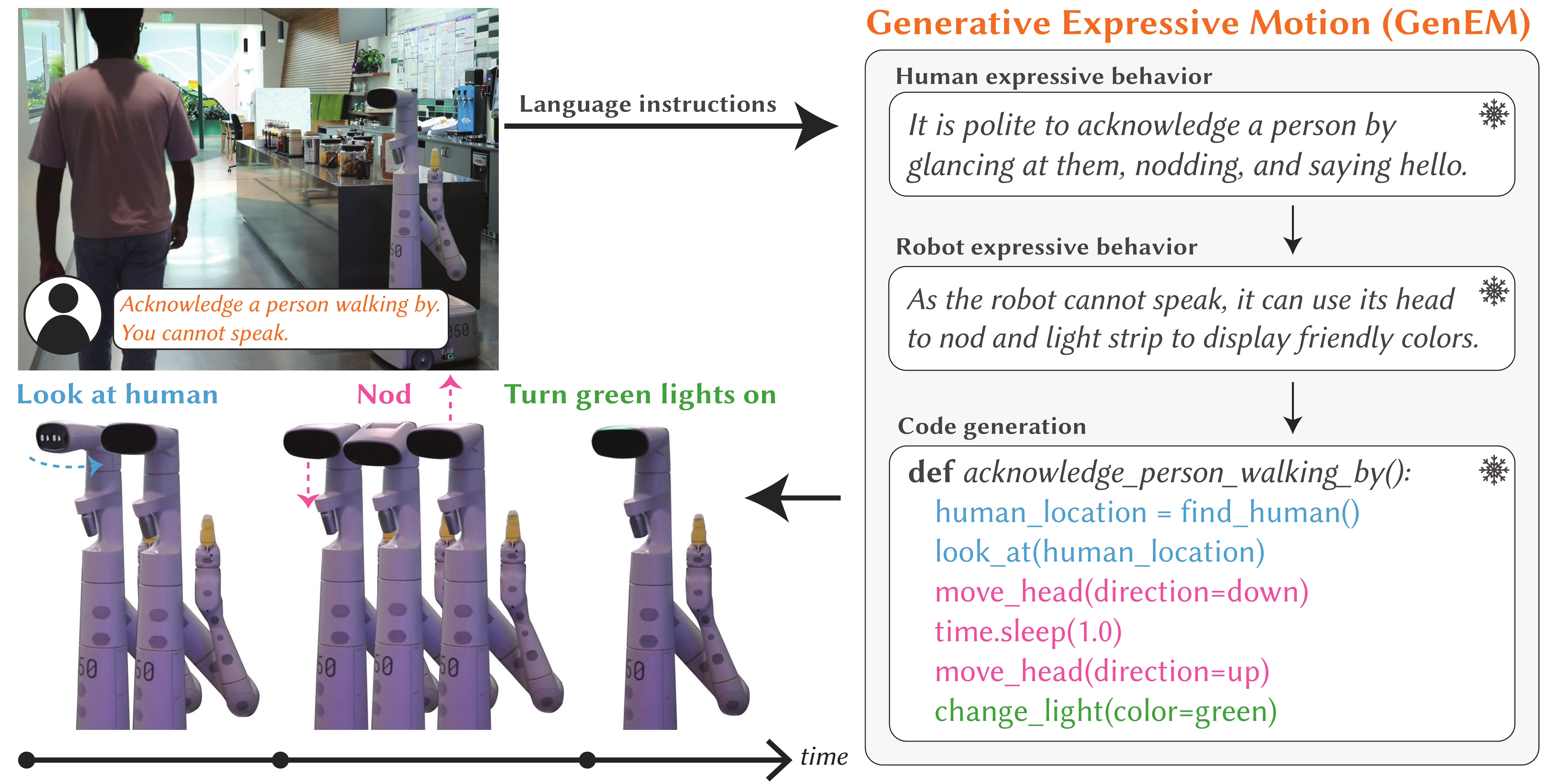}
  \vspace*{-6mm}
  \caption{We present Generative Expressive Motion (GenEM), a new approach to autonomously generate expressive robot behaviors. \abv takes a desired expressive behavior (or a social context) as language instructions, reasons about human social norms, and generates control code for a robot using pre-existing robot skills and learned expressive behaviors. Iterative feedback can quickly modify the behavior according to user preferences. Here, the * symbols denote frozen large language models.}
  \label{fig:teaser}
\end{figure}

\section{Introduction}
People employ a wide range of expressive behaviors to effectively interact with others on a daily basis. For instance, a person walking by an acquaintance may briefly glance at them and nod to acknowledge their presence. A person might apologetically say, ``excuse me!'' to squeeze through a tight hallway, where a group of people are conversing. In much the same manner, we would like robots to also demonstrate expressive behaviors when interacting with people. Robots that don't have expressive capabilities will need to re-plan their paths to avoid the crowded hallway. On the other hand, robots that have expressive capabilities might actually be able to persuade the group of people to make room for them to squeeze by, thereby improving the robot's efficiency in getting its job done.

Prior work has demonstrated the value of expressive robot behaviors, and explored approaches for generating behaviors for various purposes and contexts, including general-purpose use~\cite{desai2019geppetto}, manipulation settings, where transparency is important~\cite{kwon2018expressing}, and everyday scenarios where social norms must be observed (such as interacting with a receptionist)~\cite{porfirio2020transforming}. Approaches can be rule- or template-based~\cite{aly2013model,david2022interaction,oralbayeva2023data}, which often rely on a rigid template or a set of rules to generate behaviors. This often leads to robot behaviors that can be expressive, but do not scale to new modalities or variations of human preferences. On the other hand, data-driven techniques offer the promise of flexibility and the ability to adapt to variations. Prior work have studied data-driven techniques that generate expressive motion~\cite{suguitan2020moveae}, but these methods also have their shortcomings as they often need specialized datasets for each social interaction where a particular behavior is used (e.g., for affective robot movements~\cite{suguitan2020moveae,suguitan2019affective}).

Our goal is to enable robots to generate expressive behavior that is flexible: behaviors that can \emph{adapt} to different human preferences, and be \emph{composed} of simpler behaviors. Recent work show that large language models (LLMs) can synthesize code to control virtual~\cite{wang2023voyager} and embodied agents~\cite{liang2023code, singh2023progprompt}, help design reward functions~\cite{kwon2023reward,yu2023language}, enable social and common-sense reasoning~\cite{kwon2023toward}, or perform control and sequential decision making tasks through in-context learning~\cite{dong2022survey, min2022rethinking,mirchandani2023large} by providing a sequence of desirable inputs, and outputs in the prompt.
Our key insight is to tap into the rich social context available from LLMs to generate adaptable and composable expressive behavior. For instance, an LLM has enough context to realize that it is polite to make an eye contact when greeting someone.
In addition, LLMs enable the use of corrective language such as ``bend your arm a bit more!'' and the ability to generate motion in response to such instructions.
This makes LLMs a useful framework for autonomously generating expressive behavior that flexibly respond to and learn from human feedback in human-robot interaction settings. 



Leveraging the power and flexibility provided by LLMs, we propose a new approach, \fullname (\abv), for autonomously generating expressive robot behaviors. \abv uses few-shot prompting and takes a desired expressive behavior (or a social context) as language instructions, performs social reasoning (akin to chain-of-thought~\cite{wei2022chain}), and finally generates control code for a robot using available robot APIs. \abv can produce multimodal behaviors that utilize the robot's available affordances (e.g., speech, body movement, and other visual features such as light strips) to effectively express the robot's intent. One of the key benefits of \abv is that it responds to live human feedback -- adapting to iterative corrections and generating new expressive behaviors by composing the existing ones.

In a set of online user studies, we compared behaviors generated on a mobile robot using two variations of \abv, with and without user feedback (a non-expert in HRI behavior design), to a set of behaviors designed by a professional character animator (or the \emph{oracle animator}). We show that behaviors generated by \abv and further adapted with user feedback were positively perceived by users, and in some cases better perceived than the oracle behaviors.


In additional experiments with the mobile robot and a simulated quadruped, we show that \abv: (1) performs better than a version where language instructions are directly translated into code, (2) allows for the generation of behaviors that are agnostic to embodiment, (3) allows for the generation of composable behaviors that build on simpler expressive behaviors, and finally, (4) adapt to different types of user feedback.

%% file: Camera-Ready/Sections/Related_Work.tex
\begin{figure*}[!htb]
  \centering
  \includegraphics[width=1\linewidth]{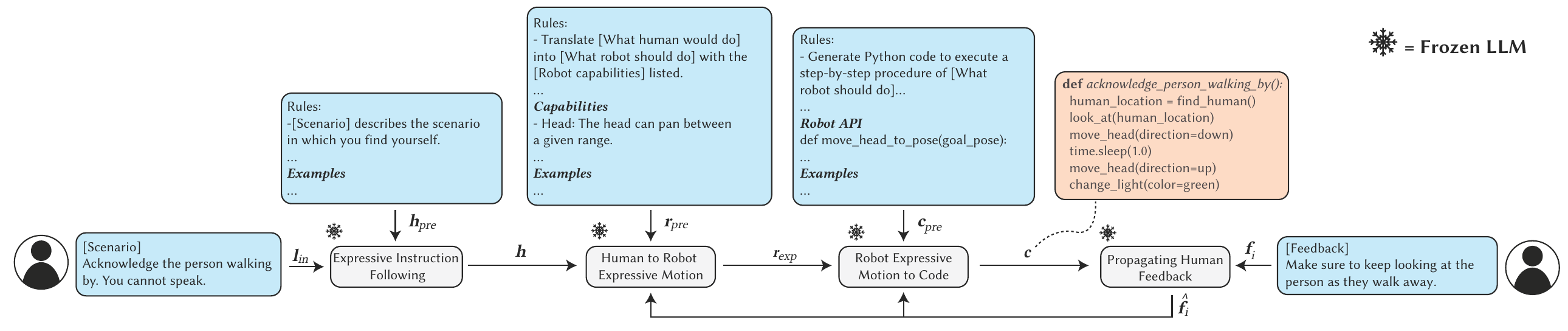}
  \vspace*{-6mm}
  \caption{\fullname. Given a language instruction $l_{in}$, the \textit{Expressive Instruction Following} module reasons about the social norms and outputs how a human might express this behavior ($h$). This is translated into a procedure for robot expressive behavior using a prompt describing the robot's pre-existing capabilities ($r_{pre}$) and any learned expressive behaviors. Then, the procedure is used to generate parametrized robot code $c$ that can be executed. The user can provide iterative feedback $f_i$ on the behavior which is processed to determine whether to re-run the robot behavior module first followed by the code generation module or just the code generation module. \emph{Note: * shown on top of all the gray modules denotes them as frozen LLMs.}}
  \label{fig:approach}
\end{figure*}

\section{Related Work}
\noindent \textbf{Expressive Behavior Generation.} Researchers have made significant efforts towards generating socially acceptable behavior for both robots and virtual humans. These can largely categorized into \textit{rule-based}, \textit{template-based}, and \textit{data-driven}~\cite{oralbayeva2023data} behavior generation approaches. We define rule-based approaches as those that require a formalized set of rules and operations (typically provided by a person) which are used to generate subsequent robot behavior.

Rule-based approaches enable behavior generation through formalized sets of rules and operations~\cite{aly2013model}. Some methods include interfaces that lets users manually specify interaction rules and logic~\cite{leonardi2019trigger, porfirio2018authoring, buchina2016design, chung2016iterative, li2020animated}. Other methods work by observing and modelling humans~\cite{kato2015may, bergstrom2008modeling, huang2012robot, huang2013repertoire}. Despite their use, rule-based approaches face several issues, including limited expressivity in the generated behavior due to the requirement of formal rules, and the reduced ability to produce multimodal behaviors as the number of modalities increases~\cite{oralbayeva2023data}. Template-based methods formulate generic templates for interaction by learning from traces of interaction data~\cite{david2022interaction, ferrarelli2018design}. Templates can translate few examples of human traces into reusable programs through program synthesis~\cite{kubota2020jessie, porfirio2019bodystorming}. Traces can be collected by observing humans interacting~\cite{porfirio2019bodystorming, porfirio2020transforming}, or through approaches such as sketching~\cite{porfirio2023sketching} or tangibles on a tabletop~\cite{porfirio2021figaro}. Overall, prior rule- and template-based methods enforce strong constraints to enable behavior generation but are limited in their expressivity. In contrast, \abv enables increased expressivity in the initial behavior generation as well as iterative improvements through live user feedback.

On the other hand, data-driven approaches produce behaviors using models trained on data. Some methods learn \textit{interaction logic} through data and use this to produce multimodal behaviors via classical machine learning methods~\cite{doering2019modeling, liu2016data, huang2014learning}. Other methods train on hand-crafted examples through generative models~\cite{suguitan2020moveae, marmpena2019generating}. For instance, predicting when to use backchanneling behaviors (i.e., providing feedback during conversation such as by nodding) has been learned through batch reinforcement learning~\cite{hussain2022training} and recurrent neural networks~\cite{murray2022learning}. Lastly, recent work has investigated how to learn cost functions for a target emotion from user feedback~\cite{zhou2018cost}, or even learn an emotive latent space to model many emotions~\cite{sripathy2022teaching}. However, these approaches are data inefficient and require specialized datasets per behavior to be generated, while \abv is able to produce a variety of expressive behaviors with a few examples through in-context learning.




\noindent \textbf{LLMs for Robot Planning and Control.}
Recent work has achieved great success by leveraging LLMs in downstream robotics tasks specifically by providing sequences of desirable input-output pairs in context~\cite{mirchandani2023large, dong2022survey, min2022rethinking}. In addition, LLMs have been used for long-horizon task planning~\cite{ahn2022can, lin2023text2motion}, and can react to environmental and human feedback~\cite{huang2023inner}. LLMs have been leveraged for designing reward functions for training reinforcement learning agents~\cite{kwon2023reward, yu2023language}. Research has also shown that LLMs can enable social and common-sense reasoning~\cite{kwon2023toward} as well as infer user preferences by summarizing interactions with humans~\cite{wu2023tidybot}. Most relevant to our approach are prior work where LLMs synthesize code to control virtual~\cite{wang2023voyager} and robotic agents~\cite{liang2023code,singh2023progprompt} by using existing APIs to compose more complex robot behavior as programs. We are also encouraged by work demonstrating that language can be used to correct robot manipulation behaviors online~\cite{cui2023no}. Taken together, we propose to leverage the rich social context available from LLMs, and their ability to adapt to user instructions, to generate expressive robot behaviors. To our knowledge, LLMs have not previously been used to generate expressive robot behaviors that adapt to user feedback.

%% file: Camera-Ready/Sections/Approach.tex
\section{\fullname}

\textbf{Problem Statement.} We aim to tackle the problem of expressive behavior generation  that is both adaptive to user feedback and composable so that more complex behaviors can build on simpler behaviors. Formally, we define being \emph{expressive} as the distance between some expert expressive trajectory that could be generated by an animator (or demonstrated) $ \tau_{\text{expert}}$ and a robot trajectory $ \tau $. $\text{dist}(\tau, \tau_{\text{expert}})$ can be any desirable distance metric between the two trajectories, e.g., dynamic time warping (DTW). \abv aims to minimize this distance $ d^* = \min \text{dist}(\tau,\tau_{\text{expert}}) $. 

Our approach (\autoref{fig:approach}) uses several LLMs in a modular fashion so that each \emph{LLM agent} plays a distinct role. Later, we demonstrate through experiments that a modular approach yields better quality of behaviors compared to an end-to-end approach. \abv takes user language instructions $ l_{in} \in L $ as input and outputs a robot policy $ \pi_\theta $, which is in the form of a parameterized code.
Human iterative feedback $ f_i \in L $ can be used to update the policy $\pi_\theta$. The policy parameters get updated one step at a time given the feedback $f_i$, where $i \in \{1,\dots, K\}$.
The policy can be instantiated from some initial state $ s_0 \in S $ to produce trajectories $ \tau = \{s_0, a_0, \dots, a_{N-1},s_N\}$ or instantiations of expressive robot behavior. Below we describe one sample iteration with human feedback $f_i$. Please refer to \textbf{Appendix A} for full prompts.


\noindent \textbf{Expressive Instruction Following.} The input to our approach is a language instruction $ l_{in} \in L$, which can either be a description of a social context where the robot needs to perform an expressive behavior by following social norms (e.g., ``A person walking by waves at you.'') \emph{or} an instruction that describing an expressive behavior to be generated (e.g., ``Nod your head''). 
The input prompt is of the form $ u = [h_{pre}, l_{in}] $ where $ h_{pre} $ is the prompt prefix that adds context about the role of the LLM and includes few-shot examples. The output of the LLM call is a string of the form $ h = [h_{cot}, h_{exp}] $ consisting of Chain-of-Thought reasoning $ h_{cot} $ ~\cite{wei2022chain} and the human expressive motion $ h_{exp} $ in response to the instruction. For example, for $ l_{in} = $ \emph{``Acknowledge a person walking by. You cannot speak.''}, the \emph{Expressive Instruction Following} module would output $ h_{exp} = $ \emph{Make eye contact with the person. Smile or nod to acknowledge their presence.} Examples of $h_{cot} $ could be: \emph{``The person is passing by and it's polite to acknowledge their presence. Since I cannot speak, I need to use non-verbal communication. A nod or a smile is a universal sign of acknowledgement.''}


\noindent \textbf{From Human Expressive Motion to Robot Expressive Motion.} In the next step, we use an LLM to translate human expressive motion $ h $ to robot expressive motion $ r $. The prompt takes the form $ u = [r_{pre}, l_{in}, h, r_{i - 1_{opt}}, \hat{f_{i - 1_{opt}}}] $ where $ r_{pre} $ is the prompt prefix setting context for the LLM, contains few-shot examples, and describes the robot's capabilities some of which are pre-defined (e.g., the ability to speak or move its head) and others which are learned from previous interactions (e.g., nodding or approaching a person). Optionally, the prompt can include the response from a previous step $ r_{i-1} $ and response to user iterative feedback from a previous step $ \hat{f_{i-1}} $. The output is of the form $ r = [r_{cot}, r_{exp}] $ consisting of the LLM's reasoning and the procedure to create expressive robot motion. An example response $ r_{exp} $ could include: \emph{``1) Use the head's pan and tilt capabilities to face the person who is walking by. 2) Use the light strip to display a pre-programmed pattern that mimics a smile or nod.''}. An example of $r_{cot} $ could be: \emph{``The robot can use its head's pan and tilt capabilities to make "eye contact" with the person. The robot can use its light strip to mimic a smile or nod.''}.

\noindent \textbf{Translating Robot Expressive Motion to Code.} In the following step, we use an LLM to translate the step-by-step procedure of how to produce expressive robot motion into executable code. We propose a skill library in a similar fashion to that of Voyager~\cite{wang2023voyager} containing existing robot skill primitives, and parametrized robot code $ \pi_\theta $ representing previously learned expressive motions. To facilitate this, the prompt encourages modular code generation by providing examples where small, reusable functions with docstrings and named arguments are used to generate more complex functions that describe an expressive behavior. To generate code, the prompt to the LLM takes the form $ u = [c_{pre}, l_{in}, h_{exp}, r_{exp,i - 1_{opt}}, c_{i - 1_{opt}}, \hat{f_{i - 1}}, \allowbreak r_{exp}]$. Here, $ c_{pre} $ provides context about its role as a code generating agent to the LLM, includes the robot's current skill library, and contains few-shot examples. Optionally, the expressive robot motion $ r_{exp, i - 1} $, and code $ c_{i - 1}$ from a previous step can be provided as well as LLM output $ \hat{f_{i - 1}} $ responding to the user feedback $ f_{i - 1} $ . The output $ c $ is parametrized robot code representing the policy $ \pi_\theta $ for the expressive behavior (see \autoref{fig:approach} for sample output). Later, the generated code can be incorporated into the robot's skill library to utilize in future expressive behavior generations.

\noindent \textbf{Propagating Human Feedback.} In the final (optional) step, we use an LLM to update the generated expressive behavior in response to human feedback $ f_{i} $ if the user is not satisfied with the generated behavior. The prompt is of the form $ u = [f_{pre}, l_{in}, r_{exp}, c, f_{i}] $, where $ f_{pre} $ provides context to LLM, and includes both the procedure for expressive robot motion $ r_{exp} $ and the generated code $ c $. The output is of the form $ f = [f_{cot}, \hat{f_{i}}] $ and includes the LLM's reasoning and the changes $ \hat{f_{i}} $ needed to improve the current expressive motion based on human feedback. The output also classifies whether the changes require an iterative call to modify the procedure for generating the robot's expressive behavior $ r $ and then translating it to code $ c $, \textit{or} just modifying the generated code $ c $. 

For example, the user could state $ f_i = $ \emph{``When you first see the person, nod at them.''}, and the output $ \hat{f_i} $ could be: \emph{``[Change: What robot should do]...As soon as the robot sees the person, it should nod at them. After nodding, the robot can use its light strip to display a pre-programmed pattern that mimics a smile or nod...''}. As an example, $ f_{cot} $ could state: \emph{`` The feedback suggests that the robot's action of acknowledging the person was not correct. This implies that the robot should nod at the person when it first sees them.''}


\begin{figure*}[!htb]
  \centering
  \includegraphics[width=1\textwidth]{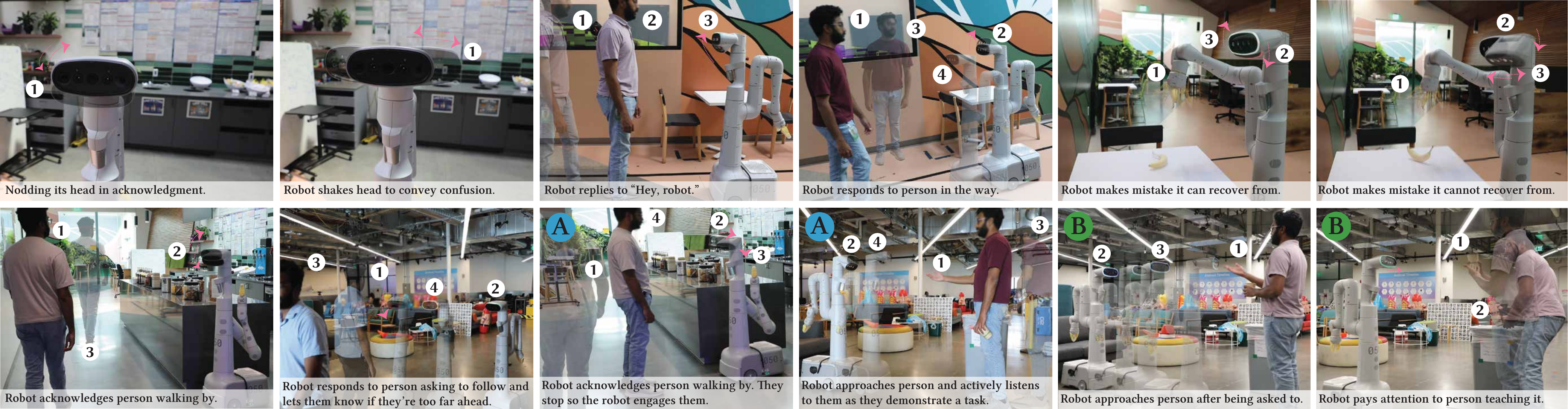}
  \vspace*{-6mm}
  \caption{Behaviors tested in the two user studies where the behaviors labelled in green denote those unique to the first study and behaviors labelled in blue denote those unique to the second study. The remaining behaviors (8) were common among the two studies.}
  \label{fig:study_behaviors}
\end{figure*}

%% file: Camera-Ready/Sections/User_Studies.tex
\section{User Studies}
We conducted two user studies to assess whether our approach, \abv, can be used to generate expressive behaviors that are perceivable by people. We generated two versions of behaviors: \emph{\genem}, and GenEM with iterative Feedback (or \emph{\genemp}). In both studies, all comparisons were made against behaviors designed by a professional animator and implemented by a software developer, which we term the \emph{\oracle}. In the \emph{first study}, our goal was to assess whether behaviors that are generated using \genem and \genemp would be perceived similarly to the behaviors created using the \oracle. In the \emph{second study}, we attempted to generate behaviors using \genem and \genemp that were similar to the behaviors created using the \oracle. Both studies aim to demonstrate that our approach is \textit{adaptable} to human feedback.


\noindent \textbf{Behaviors.} All behaviors were generated on a mobile robot platform (please see website~\footnote{\label{website}https://generative-expressive-motion.github.io/} for full clips). The robot has several capabilities that can be used to generate behaviors through existing APIs, including a head that can pan and tilt, a base that can translate, rotate, and navigate from point to point, a light strip that can display different colors and patterns, and finally, a speech module that can generate utterances and nonverbal effects. To enable the comparison of behaviors produced in the three conditions -- \oracle, \genem, and \genemp, we recorded video clips of each behavior (see \autoref{fig:study_behaviors}). To ensure consistency across conditions, behaviors in each condition were recorded in the same physical locations under similar lighting conditions. The \genem and \genemp behaviors were generated by sampling OpenAI's GPT-4 APIs for text completion~\cite{openai2023gpt4} (gpt-4-0613) with the temperature set to 0. 

\noindent \textbf{Study Procedure.} After providing informed consent, participants completed an online survey to evaluate the robot's expressive behaviors in both studies. The survey is divided into three sections (one per behavior condition) and clips within each condition randomly appeared. To minimize ordering effects, a Balanced Latin Square design (3 x 3) was used. For each behavior in each condition, participants watched an unlabeled video clip~\footref{website}, and then answered questions. All participants received remuneration after the study.

\noindent \textbf{Measures.} In both studies, participants completed a survey to assess each behavior, answering three 7-point Likert scale questions assessing their confidence on their understanding of the behavior, the difficulty in understanding what the robot is doing, and the competency of the robot's behavior. Participants also provided an open-ended response describing what behavior they \textit{believed} the robot was attempting to express.

\noindent \textbf{Analysis.} One-way repeated-measures ANOVA were performed on the data with post-hoc pairwise comparisons where there were significant differences with Bonferroni corrections applied. When reporting comparisons between conditions, we define \emph{instances} as pairwise significant conditions for at least one of the three Likert-scale questions asked about a behavior.

\subsection{Study 1: Benchmarking Generative Expressive Motion}
To determine whether our approach produces expressive behaviors that people can perceive, we conducted a within-subjects user study with thirty participants (16 women, 14 men), aged 18 to 60 (18-25: 3, 26-30: 9, 31-40: 9, 41-50: 7, 51-60: 2). One participant did not complete the entire survey and their data was omitted.

\noindent \textbf{Behaviors.} We generated ten expressive behaviors (see \autoref{fig:study_behaviors}) ranging in complexity: \emph{Nod}, shake head (\emph{Shake}), wake up (\emph{Wake}), excuse me (\emph{Excuse}), recoverable mistake (\emph{Recoverable}), unrecoverable mistake (\emph{Unrecoverable}), acknowledge person walking by (\emph{Acknowledge}), follow person (\emph{Follow}), approach person (\emph{Approach}) and pay attention to person (\emph{Attention}). The input included a one-line instruction (e.g., \emph{Respond to a person saying, ``Come here. You cannot speak.''}).

\noindent \textbf{Conditions.} The \oracle condition consisted of professionally animated behaviors that were implemented on the robot through scripting. To create the \genem behaviors, we sampled our approach five times to generate five versions of each behavior. Since the behaviors were sampled with a temperature of 0, they shared significant overlap with small variations amongst them (due to nondeterminism in GPT-4 output; please see \textbf{Appendix C} for samples generated using the same prompt). Then, six participants experienced in working with the robot were asked to rank them. The best variation for each behavior was included as part of the \genem behaviors. To generate the \genemp behaviors, we recruited one participant experienced in using the robot (but inexperienced in HRI behavior design) and asked them to provide feedback on the best rated version of each behavior. Feedback was used to iteratively modify the expressive behavior until the participant was satisfied with the result, or upon reaching the maximum number of feedback rounds (n = 10). We note that although participants rated the behaviors in the studies, the behavior generation is personalized to the user who provided the initial feedback, which may not reflect the preferences of all potential users (e.g., study participants).

\noindent \textbf{Hypotheses.} We hypothesized that the perception of the \genemp behaviors would not differ significantly from the \oracle behaviors (\textbf{H1}). We also hypothesized that the \genem behaviors would be less well-received compared to the \genemp and the \oracle behaviors (\textbf{H2}).

\begin{figure*}
  \centering
  \includegraphics[width=0.8\textwidth]{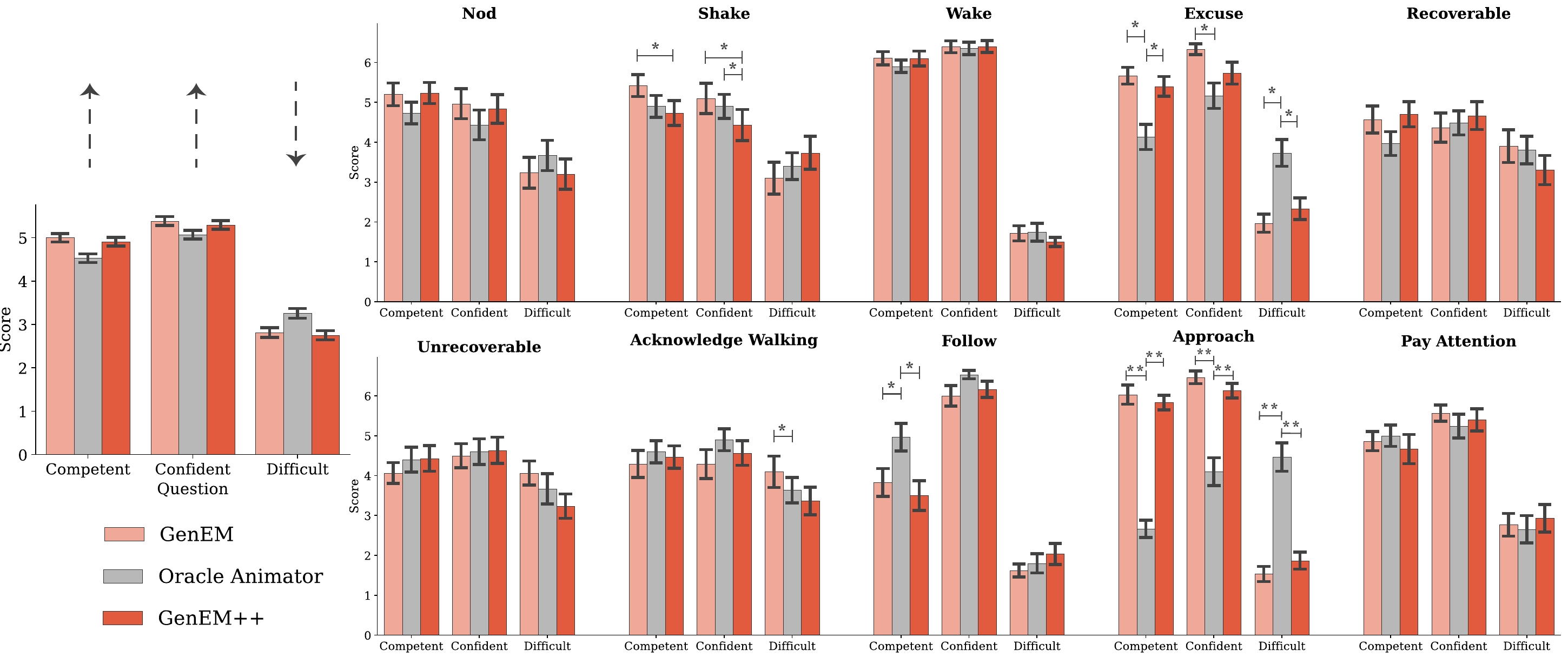}
   \vspace*{-3mm}
  \caption{Plots showing participants' survey responses to three questions about each behavior (of 10) in each condition (of 3) in the 1st user study. Bars at the top denote significant differences, where (*) denotes p<.05 and (**) denotes p<.001.  Error bars represent standard error. The first plot shows the average score for each question across conditions. The arrows reflect the direction in which better scores lie.}
  \label{fig:study1_results}
\end{figure*}

\noindent \textbf{Quantitative Findings.} \autoref{fig:study1_results} summarizes participants' responses to the survey questions for each behavior. The results show that the \genemp behaviors were worse than the \oracle behaviors in 2/10 instances (\emph{Shake} and \emph{Follow}). In contrast, the \genemp behaviors received higher scores than the \oracle behaviors in 2/10 instances (\emph{Excuse} and \emph{Approach}). Hence, \textbf{H1} is supported by our data -- the \genemp behaviors were well received and the \oracle behaviors were not significantly better received than the \genemp behaviors. 

The \genem behaviors were worse received compared to the \oracle behaviors in 2/10 instances (\emph{Acknowledge Walk} and \emph{Follow}) whereas the \genem behaviors were better received than the \oracle behaviors in 2/10 instances (\emph{Excuse} and \emph{Approach}). This was surprising because user feedback was not incorporated into the behavior generation in this condition. Besides 1/10 instances (\emph{Shake}), there were no significant differences in the perceptions of the \genem and \genemp behaviors. Hence, we did not find support for \textbf{H2}. We performed equivalence tests (equivalence bound: +/- 0.5 Likert points) but did not find any sets of behaviors to be equivalent. Overall, the results support the finding that \abv (even with an untrained user providing feedback) produces expressive robot behaviors that users found to be competent and easy to understand.





\begin{figure*}
  \centering
  \includegraphics[width=0.8\textwidth]{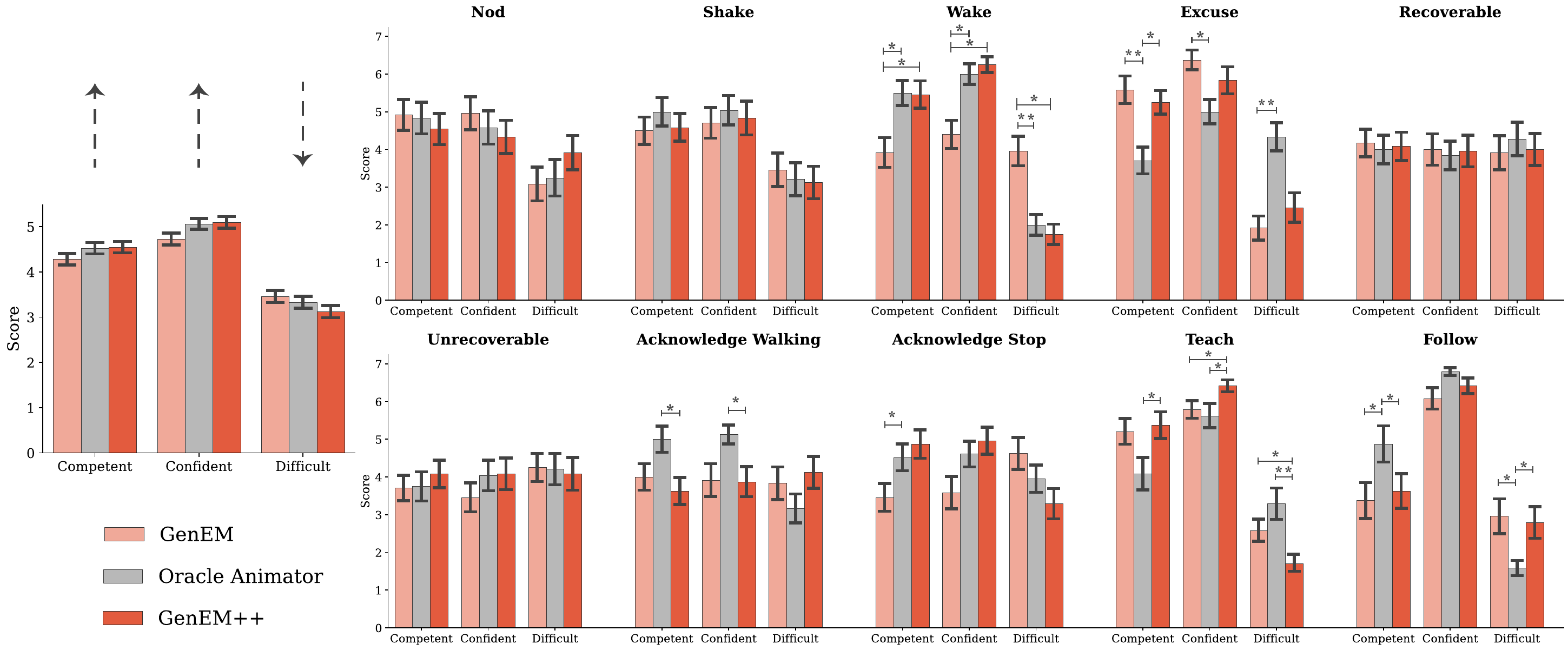}
  \vspace*{-3mm}
  \caption{Plots showing participants’ survey responses to three questions about each behavior (of 10) in each condition (of 3) in the 2nd user study. Bars at the top denote significant differences, where (*) denotes p<.05 and (**) denotes p<.001. Error bars represent standard error. The first plot shows the average score for each question across conditions. The arrows reflect the direction in which better scores lie.}
  \label{fig:study2_results}
\end{figure*}

\subsection{Study 2: Mimicking the Oracle Animator} 
We conducted an additional within-subjects user study with twenty four participants (21 men, 2 women, 1 prefer not to say), aged 18-60 (18-25: 4, 26-30: 3, 31-40: 12, 41-50: 4, 51-60: 1) to assess whether using \abv to generate behaviors that resembled the \oracle would be perceived differently. One participant did not complete the entire survey and their data was omitted.

\noindent \textbf{Behaviors.} We generated ten expressive behaviors ranging in complexity, with eight overlapping~\footnote{Some behaviors in the second study differ from the first study as they are too complex to express as a single line instruction which we maintained for consistency in the first study. Instead, in the first study, these complex behaviors were broken down into simpler behaviors (e.g., teaching is equivalent to approaching and paying attention).} behaviors from the first study (see \autoref{fig:study_behaviors}): nod (\emph{Nod}), shake head (\emph{Shake}), wake up (\emph{Wake}), excuse me (\emph{Excuse}), recoverable mistake (\emph{Recoverable}), unrecoverable mistake (\emph{Unrecoverable}), acknowledge person walking by (\emph{Acknowledge Walking}), acknowledge person stopping by (\emph{Acknowledge Stop}), follow person (\emph{Follow}), and teaching session (\emph{Teach}). Behaviors that were different from the first study were chosen to add further complexity -- e.g., longer single-turn interactions such as \textit{teaching}, that started with a person walking up a robot, teaching it a lesson, and lastly the robot acknowledging that it understood the person's instructions. Unlike in the first study, the prompts were more varied and sometimes included additional descriptions such as for the more complex behaviors (see \textbf{Appendix B} for full prompts for each behavior). To generate each \genem behavior, we sampled our approach ten times after which an experimenter selected the version that appeared most similar to the equivalent \oracle behavior when deployed on the robot. To create each \genemp behavior, an experimenter refined the \genem behavior through iterative feedback until it appeared similar to the equivalent \oracle behavior or after exceeding the maximum number of feedback rounds (n = 10)~\footref{website}. 

\noindent \textbf{Hypotheses.} We hypothesized that user perceptions of the \genemp behaviors would not significantly differ when compared to the \oracle behaviors (\textbf{H3}). We also suppose that the behaviors in the \genem condition would be perceived as worse than the \genemp and \oracle behaviors (\textbf{H4}).

\noindent \textbf{Quantitative Findings.} The results of the study are summarized in \autoref{fig:study2_results}. They show that the \genemp behaviors were worse received than the \oracle behaviors in 2/10 instances (\emph{Acknowledge  Walk} and \emph{Follow}) whereas the \genemp behaviors were more positively received than the \oracle in 2/10 instances (\emph{Excuse} and \emph{Teach}). Hence, our hypothesis is supported by the data \textbf{(H3)} -- the \genemp behaviors well received and the \oracle behaviors were not significantly better perceived. When comparing the \oracle behaviors and \genem behaviors, there were 4/10 instances where the \genem behaviors were worse received (\emph{Wake}, \emph{Acknowledge Walk}, \emph{Acknowledge Stop}, and \emph{Follow}), and 1/10 instances where the \genem behaviors were more positively rated (\emph{Excuse}). As with the first study, it is somewhat surprising that the \genem behaviors were better received than the baselines in one instance; although they resemble them, they do not capture all the nuances present in the \oracle behaviors since user feedback is not provided. Lastly, the \genem behaviors were rated worse than the \genemp behaviors in 2/10 instances (\emph{Wake} and \emph{Teach}) whereas there were 0/10 instances where the reverse was true. Hence, we did not find support for the last hypothesis \textbf{(H4)}. Upon performing equivalence tests (equivalence bound: +/- 0.5 Likert points), we did not find any sets of behaviors to be equivalent. Overall, the findings suggest that expressive robot behaviors produced using our approach (with user feedback) were found competent and easy to understand by users.

%% file: Camera-Ready/Sections/Experiments.tex
\section{Experiments}
\begin{figure}
  \centering
  \includegraphics[width=1\linewidth]{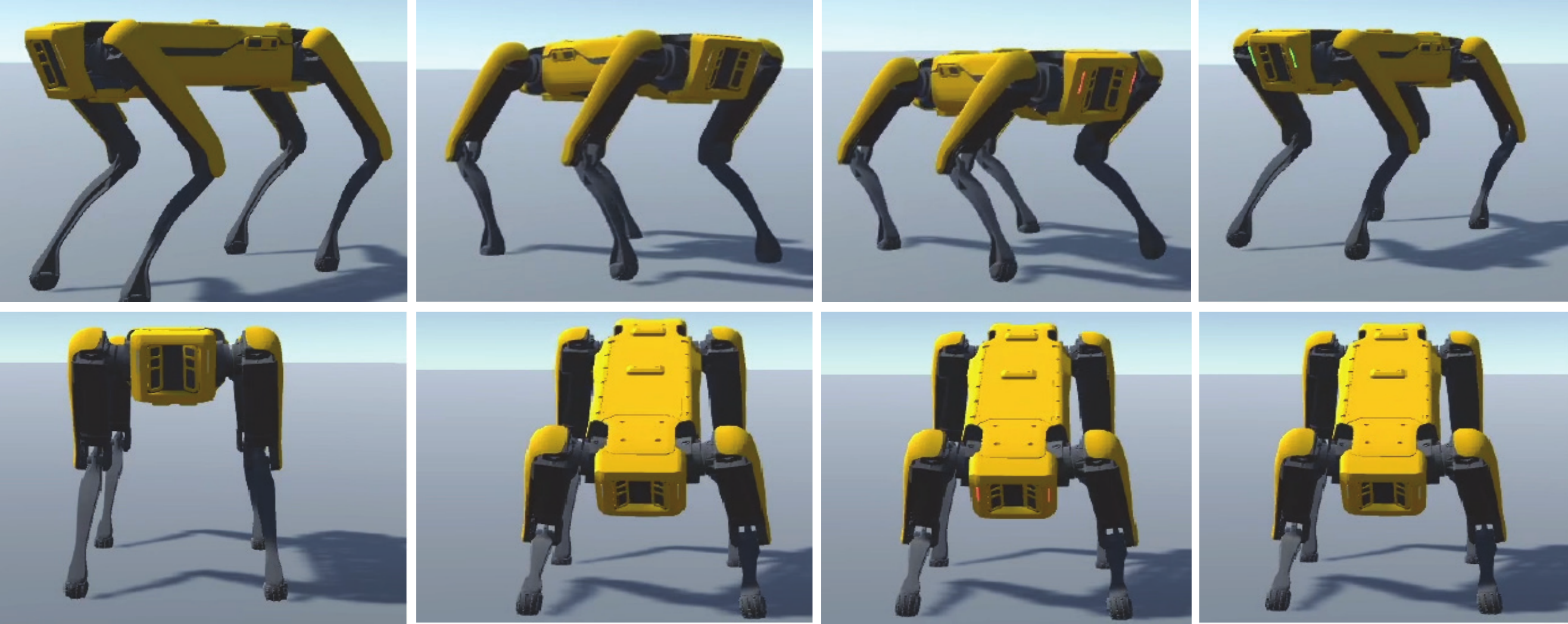}
  \vspace*{-6mm}
  \caption{Quadruped simulated in Gazebo performing the \emph{Recoverable mistake} behavior (top) and \emph{Unrecoverable mistake} (bottom) generated by GenEM prior to feedback. After making a recoverable mistake, the robot demonstrates it made a mistake by turning away, lowering its legs, and flashing red lights to convey regret but then returns to its initial position and flashes a green light. In contrast, an unrecoverable mistake causes the robot to lower its height, display red lights for a brief period, and bow forwards and maintains this pose.}
  \label{fig:quad}
\end{figure}

We conducted a set of experiments to carefully study different aspects of \abv. This includes ablations to understand the impact of our prompting structure and the modular calls to different LLMs versus an end-to-end approach. Further, through an experiment, we demonstrate that \abv can produce modular and composable behaviors, i.e., behaviors that build on top of each other. The behaviors were generated by sampling OpenAI's GPT-4 APIs for text completion~\cite{openai2023gpt4} (gpt-4-0613) with the temperature set to 0. In addition to our user study and experiments on the mobile manipulator, we conducted further experiments using a quadruped simulated in Gazebo/Unity via ROS (see~\autoref{fig:quad}).

\begin{table}
\begin{tabular}{lcccc}
\toprule
\multicolumn{1}{c}{} & \multicolumn{2}{c}{\textbf{GenEM}}  & \multicolumn{2}{c}{\textbf{Ablated}} \\
                     & \textit{Execution} & \textit{Norms} & \textit{Execution}  & \textit{Norms} \\ \midrule
Nod                  & 5                  & 0              & 5                   & 2              \\
Shake                & 5                  & 0              & 5                   & 2              \\
Wake                 & \textbf{4}         & 2              & 3                   & 0              \\
Excuse               & \textbf{5}         & 3              & 0                   & -              \\
Recoverable          & 3                  & 0              & \textbf{5}          & 1              \\
Unrecoverable        & 5                  & 0              & 5                   & 0              \\
Acknowledge          & 5                  & 1              & 5                   & 0              \\
Follow               & \textbf{3}         & 1              & 0                   & -              \\
Approach             & 5                  & 1              & 5                   & 3              \\
Attention            & \textbf{4}         & 0              & 1                   & 0              \\
\bottomrule
\end{tabular}
\caption{Ablations on the mobile robot platform showing the successful attempts of behavior generation when sampling each prompt five times to compare our approach (without feedback) against a variation without the \emph{Expressive Instruction Following} module and subsequently the module translating human expressive motion to robot expressive motion. The \emph{Execuution} column indicates the number of successful attempts (/5). The \emph{Norms} column indicates the number of attempts where social norms were not appropriately followed (coded by the experimenter).}
\label{tab:ablations}
\vspace{-7mm}
\end{table}

\begin{table}
\begin{tabular}{lcc}
             \toprule
              & \textit{Execution} & \textit{Norms} \\ \midrule
Nod           & 5                  & 0     \\
Shake         & 5                  & 0              \\
Wake          & 5                  & 0              \\
Excuse        & 3                  & 0              \\
Recoverable   & 5                  & 2              \\
Unrecoverable & 4                  & 0              \\
Acknowledge   & 4                  & 1              \\
Follow        & 2                  & 2              \\
Approach      & 5                  & 5              \\
Attention     & 1                  & 0              \\
\bottomrule
\end{tabular}
\caption{Behaviors generated on the quadruped in simulation showing successful attempts of behavior generation when sampling each prompt five times. The \emph{Execution} column indicates the number of successful attempts (/5). The \emph{Norms} column indicates the number of attempts where social norms were not properly observed (coded by the experimenter).}
\label{tab:xembodiment}
\vspace{-7mm}
\end{table}

\noindent\textbf{Ablations.} We performed ablations to compare \abv to an end-to-end approach that takes language instructions and makes one call to an LLM to generate an expressive behavior. The ablations were performed using existing APIs for the mobile robot. The behaviors examined were identical to the first user study along with the prompts. Each prompt was sampled five times to generate behaviors and executed on the robot to verify \textit{correctness}. Further, an experimenter examined the code to check whether the behavior code incorporated reasoning to account for human social norms. The results for code correctness and social norm appropriateness are shown in~\autoref{tab:ablations}. Overall, our approach produced higher success rates compared to the ablated variation where no successful runs were generated for 2 behaviors -- \emph{Excuse} and \emph{Follow}. For the \emph{Excuse} behavior, the robot must check the user's distance and signal to a person that they are in its way. However, for the ablated variation, the distance was never checked in the attempts. For the \emph{Follow} behavior, the code called functions that were not previously defined, and used the wrong input parameter type when calling robot APIs, resulting in zero successful attempts. Further, nearly all generated functions were missing docstrings and named arguments, which could make it difficult to use them in a modular fashion for more complex behaviors (despite providing few-shot code examples). 

We qualitatively observed that behaviors generated by \abv reflected social norms, particularly for more complex behaviors, and looked similar for simpler behaviors. For instance, the \emph{Excuse} behavior generated by \abv used the speech module to say, \emph{``Excuse me''}. For the \emph{Attention} behavior, the ablated variations looked at the person, turned on the light strip, and then turned it off, whereas the \abv variations also incorporated periodic nodding to mimic ``active listening''. For the \emph{Approach} behavior, the \abv variations always incorporated a nod before moving towards the person while the ablated variations never used nodding; instead lights were used in two instances.

\noindent\textbf{Cross-Embodiment Behavior Generation.} We sampled the same prompts in the first user study five times per behavior using API for a simulated Spot robot. The results, summarized in \autoref{tab:xembodiment}, show that we were able to generate most expressive behaviors using the same prompts using a different robot platform with its own affordances and APIs. However, some generated behaviors such as \emph{Approach} included variations where the robot navigated to the human's location instead of a safe distance near them, which would be considered a social norm mismatch (possibly due to the lack of a distance threshold parameter in the translate API), while some did not account for the human (e.g., the robot rotating an arbitrary angle instead of towards the human for \textit{Attention}). Overall, the success rates hint at the generality of our approach to differing robot embodiments.

\begin{table}
\begin{tabular}{lccccc}
                 & \textbf{}                                                      & \multicolumn{1}{l}{}                                               & \multicolumn{1}{l}{}                                           &                                                               & \multicolumn{1}{l}{}                                                            \\ \toprule
                 & \textit{\begin{tabular}[c]{@{}c@{}}Eye\\ contact\end{tabular}} & \textit{\begin{tabular}[c]{@{}c@{}}Blinking\\ lights\end{tabular}} & \textit{\begin{tabular}[c]{@{}c@{}}Look\\ around\end{tabular}} & \textit{\begin{tabular}[c]{@{}c@{}}Shake\\ head\end{tabular}} & \multicolumn{1}{l}{\textit{\begin{tabular}[c]{@{}l@{}}Nod\\ head\end{tabular}}} \\ \midrule
Acknowledge Walk           & 5                                                              & -                                                                  & -                                                              & \textit{-}                                                    & 5                                                                               \\
Approach         & 4                                                              & 5                                                                  & -                                                              & -                                                             & 0                                                                               \\
Confusion & -                                                              & 4                                                                  & 1                                                              & 5                                                             & -                                                                                \\   
\bottomrule 
\end{tabular}
\caption{Number of times (out of 5 attempts) where previously-learned behaviors (columns) are used when composing new behaviors (rows) using GenEM. Dashes indicate that the given learned behavior API is not provided when prompting the creation of the new behavior.}
\label{tab:composability}
\vspace{-7mm}
\end{table}

\noindent\textbf{Composing Complex Expressive Behaviors.} In the user studies, all behaviors were generated from scratch using few-shot examples and existing robot APIs. We attempted to generate more complex behaviors using a set of learned expressive behaviors from previous interactions --- these skills (represented as functions with docstrings) were appended to the prompts describing the robot's capabilities (step 2 of our approach) as well as the robot's API (step 3 of our approach). The learned behaviors used in the prompt were: \emph{nodding}, \emph{making eye contact}, \emph{blinking the light strip}, \emph{looking around}, and \emph{shaking}. We prompted \abv to generate three behaviors, varying in complexity: \emph{Acknowledge Walk}, \emph{Approach}, and expressing confusion (\emph{Confusion}). All of these behaviors were generated on the quadruped without providing feedback, using instructions that contained a single line description of the desired behavior. We sampled \abv five times to assess the frequency with which learned behaviors would be included in the outputted program. To assess success, an experimenter checked whether the generated code utilized a combination of robot APIs and learned APIs (see \autoref{tab:composability}). For the approach behavior, it was surprising to note that the \textit{nod head} behavior was never utilized whereas blinking lights were always used. For expressing confusion, it was surprising that 4/5 instances generated code for looking around, but only 1/5 instances used the existing \textit{looking around} behavior.

\begin{table}
\begin{tabular}{lcccc}
                 & \textbf{}                                                          & \multicolumn{1}{l}{}                                             & \multicolumn{1}{l}{}                                             &                                                                       \\ \toprule
                 & \textit{\begin{tabular}[c]{@{}c@{}}Insert \\ actions\end{tabular}} & \textit{\begin{tabular}[c]{@{}c@{}}Swap \\ actions\end{tabular}} & \textit{\begin{tabular}[c]{@{}c@{}}Loop \\ actions\end{tabular}} & \textit{\begin{tabular}[c]{@{}c@{}}Remove \\ capability\end{tabular}} \\ \midrule
Excuse           & 4                                                         & 5                                                                & 5                                                                & 5                                                            \\
Approach         & 4                                                                  & 5                                                                & 5                                                                & 3                                                                     \\
Acknowledge Stop & 5                                                                  & 5                                                                & 4                                                                & 3                 \\            \bottomrule
\end{tabular}
\caption{Success rates (out of 5 attempts) when providing different types of feedback to behaviors generated using \abv, where: \emph{Insert actions} request a new action be added ahead of other actions, \emph{Swap actions} request to swap the order of existing actions, \emph{Loop actions} request to add loops to repeat actions, and \textit{Remove capability} requests to swap an existing action with an alternate one. }
\label{tab:feedback}
\vspace{-7mm}
\end{table}

\noindent\textbf{Adaptability to Human Feedback.} In the user studies, feedback had some effect on the perception of the generated behaviors. Further, we qualitatively observed that feedback could steer the behavior generation in different ways. We studied this in an experiment where we generated three behaviors from the two prior studies: \emph{Excuse}, \emph{Approach}, and \emph{Acknowledge Stop}. Each behavior was generated using a single-line description as before, and without any learned robot APIs. We attempted to modify the generated behavior through four types of feedback: (1) adding an action and enforcing that it must occur before another action, (2) swapping the order of the actions, (3) making a behavior repeat itself (loops), and (4) removing an existing capability without providing an alternative (e.g., removing the light strip as a capability after producing a behavior that uses the light strip). Overall, the results (see~\autoref{tab:feedback}) suggest that it is possible to modify the behavior according to the type of feedback provided, though removing capabilities lead to calling undefined functions more often.

%% file: Camera-Ready/Sections/Discussion.tex
\section{Discussion}
\textbf{Summary.} In this work, we proposed an approach, \abv, to generate and modify expressive robot motions using large language models by translating user language instructions to robot code. Through user studies and experiments, we have shown that our framework can quickly produce expressive behaviors by way of in-context learning and few-shot prompting. This reduces the need for curated datasets to generate specific robot behaviors or carefully crafted rules as in prior work. In the user studies, we demonstrated that participants found the behaviors generated using \abv with user feedback competent and easy to understand, and in some cases perceived significantly more positively than the behaviors created by an expert animator. We have also shown that our approach is \emph{adaptable} to varying types of user feedback, and that more complex behaviors can be \emph{composed} by combining simpler, learned behaviors. Together, they form the basis for the rapid creation of expressive robot behaviors conditioned on human preferences.

\noindent\textbf{Limitations and Future Work.} Despite the promise of our approach, there are a few shortcomings. Our user studies were conducted online through recorded video clips, and although this is a valid methodology~\cite{hoffman2014designing, takayama2011expressing}, it may not reflect how participants would react when in the physical proximity of the robot~\cite{woods2006comparing}. Hence, further studies involving interactions with the robot should be pursued. Some inherent limitations of current LLMs should be noted, including small context windows and the necessity for text input. 

In our work, we only evaluate single-turn behaviors (e.g., acknowledging a passerby), but there are opportunities to generate behaviors that are multi-turn and involve back-and-forth interaction between the human and the robot. Future work should also explore generating motion with a larger action space such as by including the manipulator and gripper. Although we have shown that our approach can adapt to user feedback and their preferences, there is currently no mechanism to learn user preferences over a longer period. In reality, we expect that users will exhibit individual differences in their preferences about the behaviors they expect robots to demonstrate in a given situation. Hence, learning preferences in-context~\cite{wu2023tidybot} may be a powerful mechanism to refine expressive behaviors.



Despite these limitations, we believe our approach presents a flexible framework for generating adaptable and composable expressive motion through the power of large language models. We hope that this inspires future efforts towards expressive behavior generation for robots to more effectively interact with people.


%% file: Camera-Ready/Sections/Acknowledgements.tex
\begin{acks}
We thank Doug Dooley for providing animations for the baseline robot behaviors, and Edward Lee for helpful discussions on the system. We thank Rishi Krishnan, Diego Reyes, Sphurti More, April Zitkovich, and Rosario Jauregui for their help with robot access and troubleshooting, and Justice Carbajal, Jodilyn Peralta, and Jonathan Vela for providing support with video recording. Lastly, we thank Ben Jyenis and the UX research team for coordinating the user studies and data collection efforts.
\end{acks}